\documentclass[aip,amsmath,amssymb]{revtex4-2}
\newtheorem{Theorem}{Theorem}
\newtheorem{Proposition}[Theorem]{Proposition}
\usepackage{graphicx}

\begin{document}

\title{Two new parameters for the ordinal analysis of images} 

\author{Christoph Bandt}
 \email{bandt@uni-greifswald.de.}
\affiliation{Institute of Mathematics, University of Greifswald, Greifswald, Germany}

\author{Katharina Wittfeld}
\email{katharina.wittfeld@uni-greifswald.de}
\affiliation{German Center for Neurodegenerative Diseases (DZNE), Site Rostock/ Greifswald, Greifswald, Germany}
\affiliation{Department of Psychiatry and Psychotherapy, University Medicine Greifswald, Greifswald, Germany}

\date{\today}

\begin{abstract}
Local patterns play an important role in statistical physics as well as in image processing.  Two-dimensional ordinal patterns were studied by Ribeiro et al. who determined permutation entropy and complexity in order to classify paintings and images of liquid crystals. Here we find that the 2 by 2 patterns of neighboring pixels come in three types. The statistics of these types, expressed by two parameters, contains the relevant information to describe and distinguish textures. The parameters are most stable and informative for isotropic structures.
\end{abstract}


\maketitle  

\begin{quotation}
Local correlations and transition probabilities have a long history in statistical physics. Around 2000, physicists used local entropy and complexity to describe pattern formation \cite{Andrienko2000, Feldman2003} while computer scientists became interested in local pattern statistics of textures \cite{Pieti2011}. On ordinary level, the entropy-complexity plane was introduced by Rosso et al. \cite{Rosso2007} for one-dimensional systems, and extended to two-dimensional patterns in images by Ribeiro et al. \cite{Ribeiro2012}. Starting from the observation that $2\times 2$ patterns come in three types, this paper develops a different approach. Two parameters expressing the frequency of types define smoothness and curve structure of an image. It is shown that the parameters consistently describe textures and are well-suited to distinguish different structures.
\end{quotation}

\section{\label{intro}Introduction}
Ten years ago, Ribeiro and colleagues~\cite{Ribeiro2012} started the study of ordinal patterns in images.  Rosso et al. \cite{Rosso2007} had introduced the entropy-complexity plane to distinguish chaos and noise in one-dimensional signals. This physical background was appropriate for the study of phase transitions of liquid crystals and Ising systems in two dimensions.  The methodology was applied by Zunino and Ribeiro \cite{Zunino2016} to the study of textures in images by two-dimensional ordinal patterns on different scales. In the sequel, Sigaki et al.~\cite{Sigaki2019} predicted physical properties of liquid crystals and studied historical artwork, observing distinct clusters of artistic styles.~\cite{Sigaki2018}\  Brazhe \cite{Brazhe2018} introduced a multiscale algorithm, and Azami et al. \cite{Azami2019} suggested a new ordinal entropy concept. Pessa and Ribeiro~\cite{Pessa2020} investigated transition probabilities between neighboring ordinal patterns and corresponding networks.

There is a huge demand for fast methods evaluating textures.  An unbelievable amount of image data is produced each day, ranging from microscopic pictures in cancer and virus detection to multispectral satellite images. They have to be screened automatically to fix regions where valuable information could be. Computer must find similar phenomena and do some classification before men and women will see the picture. There are many tools for studying local structure \cite{Pieti2011}. Ordinal parameters like permutation entropy \cite{Bandt2002a} have shown to be fast, simple and robust in one dimension. It is tempting to transfer them to two dimensions where quadratic data size requires just these properties. 

Here we study $2\times 2$ ordinal patterns in images and introduce two new parameters which we call smoothness $\tau$ and curve structure $\kappa .$ The basic observation is that the 24 patterns can be divided into three types. At least for isotropic textures, the distinction between the three types seems to be the essential information contained in the frequency distribution of the 24 ordinal patterns. Taking two contrasts of frequencies, similar to the one-dimensional case \cite{Bandt2022}, images can be represented in the $\tau$-$\kappa$-plane, similar to the entropy-complexity plane.

In Section \ref{methods} we give the basic definitions and comment on interpretation, calculation and relevance of our parameters. There is also a theoretical discussion concerning statistical dependence of $\tau$ and $\kappa .$  In Section \ref{appli}, we apply our method to the Kylberg Sintorn rotation database  \cite{Kylberg2016, KSdatabase, Kylberg2014}    which contains 900 unrotated and rotated samples of 25 textures.  The results are very promising. Our parameters distinguish many of the textures, show very small variance within structures and are rotation-invariant for istropic textures.  We also apply the method to some photos and fractal surfaces.

\section{\label{methods} Concepts}

\subsection{Two-dimensional ordinal patterns}

The definition of ordinal $2\times 2$ patterns was introduced by Ribeiro and colleagues in 2012 \cite{Ribeiro2012}. We slightly change the notation of permutations, using rank numbers which are directly linked to the spatial visualization of the patterns (cf. \cite{Bandt2022}, section 2). Note that the entries of a monochrome image matrix are real numbers, often integers between 0 and 255, which represent light intensities or shades of gray (0=black, 255=white). There are other applications where the values in a matrix represent certain physical quantities.

\begin{figure}
\includegraphics[width=0.8\textwidth]{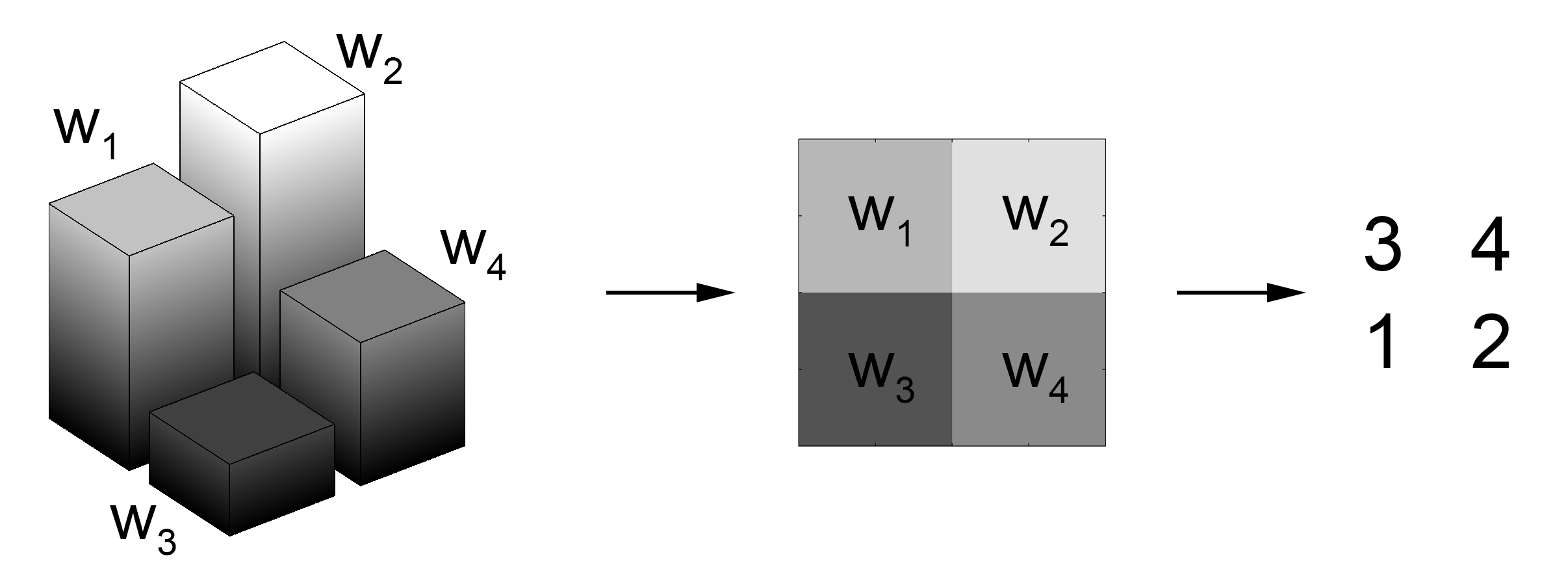}
\caption{\label{fig:ex_ordPat2D} An ordinal pattern based on the four values $w_k$ symbolized as heights on the left and as gray tone values in the middle. The assigned ordinal pattern is $\pi=\left(\begin{smallmatrix}3&4\\1&2\\\end{smallmatrix}\right)$.}
\end{figure}

\textbf{Definition.} Let $X=(x_{mn})_{m=1,2,\ldots,M\ n=1,2,\ldots,N}$ be a data matrix. The local $2\times 2$ pattern at position $(m,n)$ is the matrix
\begin{equation}
\left(\begin{matrix}
x_{m\,n}&x_{m\,n+1}\\
x_{m+1\,n}&x_{m+1\,n+1}
\end{matrix}\right) = \left(\begin{matrix}
{w_1}&{w_2}\\
{w_3}&{w_4}
\end{matrix}\right) \ . \label{locpat}
\end{equation}
We now replace the values $w_k$ by their rank numbers 1, 2, 3, 4 within this small matrix: 1 denotes the minimum, 4 denotes the maximum. In general  $r_k$ denotes the number of $w_j$ which fulfil $w_j \le w_k.$
The \textbf{$2\times 2$ ordinal pattern} $\pi_{mn}$ at position $(m,n)$ is defined as 
\[ \pi_{mn}=\left(\begin{matrix}
r_1&r_2\\
r_3&r_4 \end{matrix}\right) \ . \]
An example of the construction of an $2\times 2$ ordinal pattern is shown in Figure~\ref{fig:ex_ordPat2D}. The smallest of the values is $w_3$, which is pictured with the smallest height on the left and the darkest gray tone. On the right-hand side of the figure are the ranks which form the ordinal pattern.

Instead of the matrix \eqref{locpat} of immediately neighboring values, one can also take 
the values $x_{m\; n}$, $x_{m\;n+d}$, $x_{m+d\;n}$, and $x_{m+d\;n+d}$ which are $d$ steps apart. The resulting matrix of ranks will then be called ordinal pattern at position $(m,n)$ with delay $d.$ Our initial definition of ordinal pattern is the case $d=1.$ Clearly $d$ is a scale parameter, we may think of a zoom factor. 
In the study of time series in one dimension, $d$ is called embedding delay. Zunino and Ribeiro \cite{Zunino2016} studied different delays for the $x$- and $y$-direction.

As usual in the study of ordinal patterns, we exclude equality of values which are compared, and enforce this assumption by adding a tiny white noise to the data matrix~$X.$ 
When the number of gray tones is small, however, ties among neighbouring values will be quite frequent.  For the types defined below, a better treatment of ties seems possible. This is work in progress.

\subsection{Three types of $2\times 2$ patterns}
\begin{figure}
\includegraphics[width=0.8\textwidth]{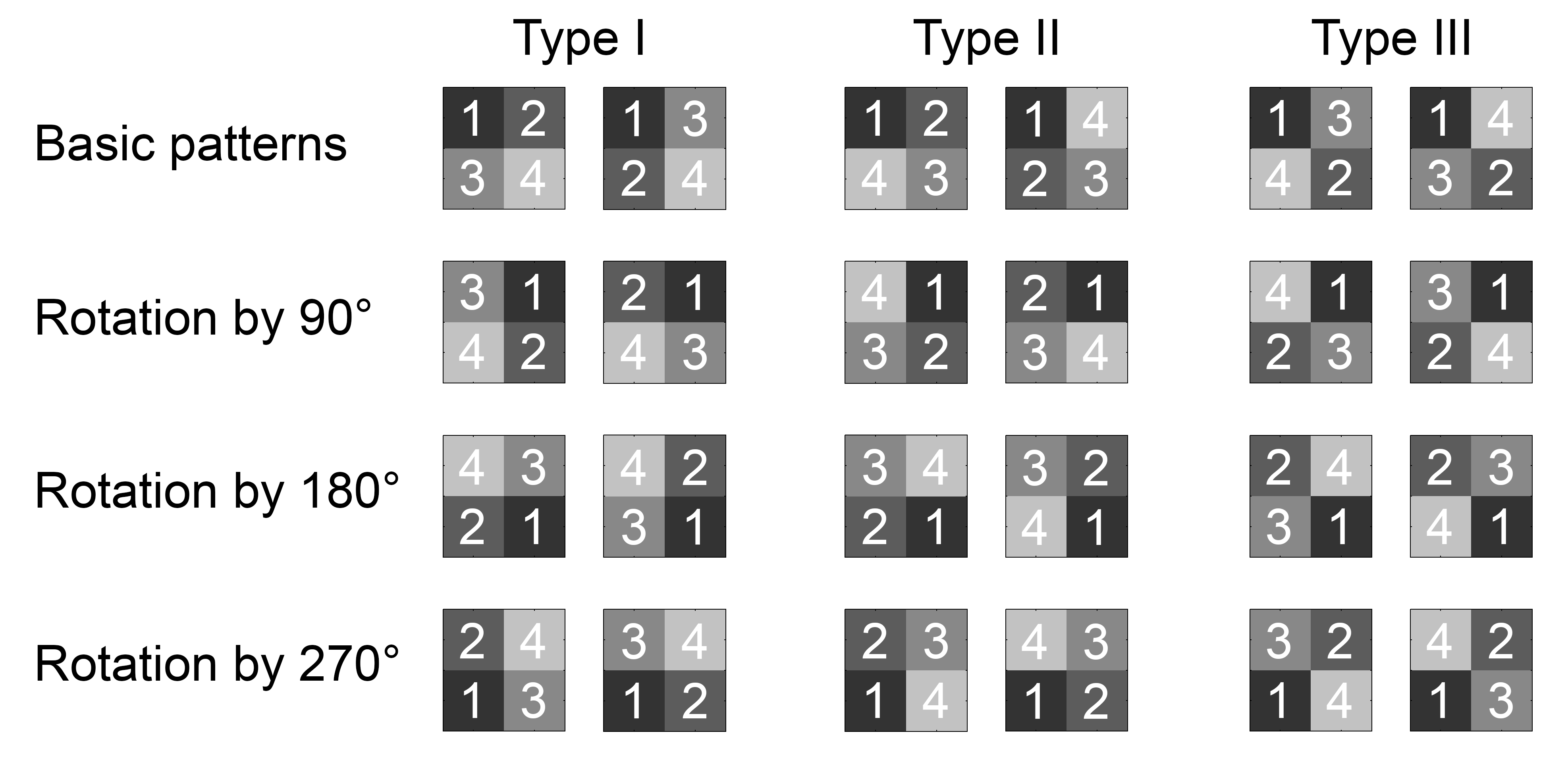}
\caption{\label{fig:ordPat2D_overview} 24 ordinal patterns grouped into three different types.}
\end{figure}

There are $4!\ =24$ different $2\times 2$ ordinal patterns. Let us arrange them according to shape characteristics. Fixing rank 1 at the upper left corner, we obtain six basic ordinal patterns, shown in the first row of Figure~\ref{fig:ordPat2D_overview}. They form three pairs of patterns according to the ranks which are diagonally opposite of each other. Rotating the six basic patterns by $90^{\circ}$, $180^{\circ}$, and $270^{\circ}$, we derive the other 18 patterns.
Following this procedure, the 24 ordinal patterns can naturally be grouped in three sets of eight patterns each which we would like to denote with “type I”, “type II”, and “type III” (Figure~\ref{fig:ordPat2D_overview}). 

\begin{figure}
\includegraphics[width=0.8\textwidth]{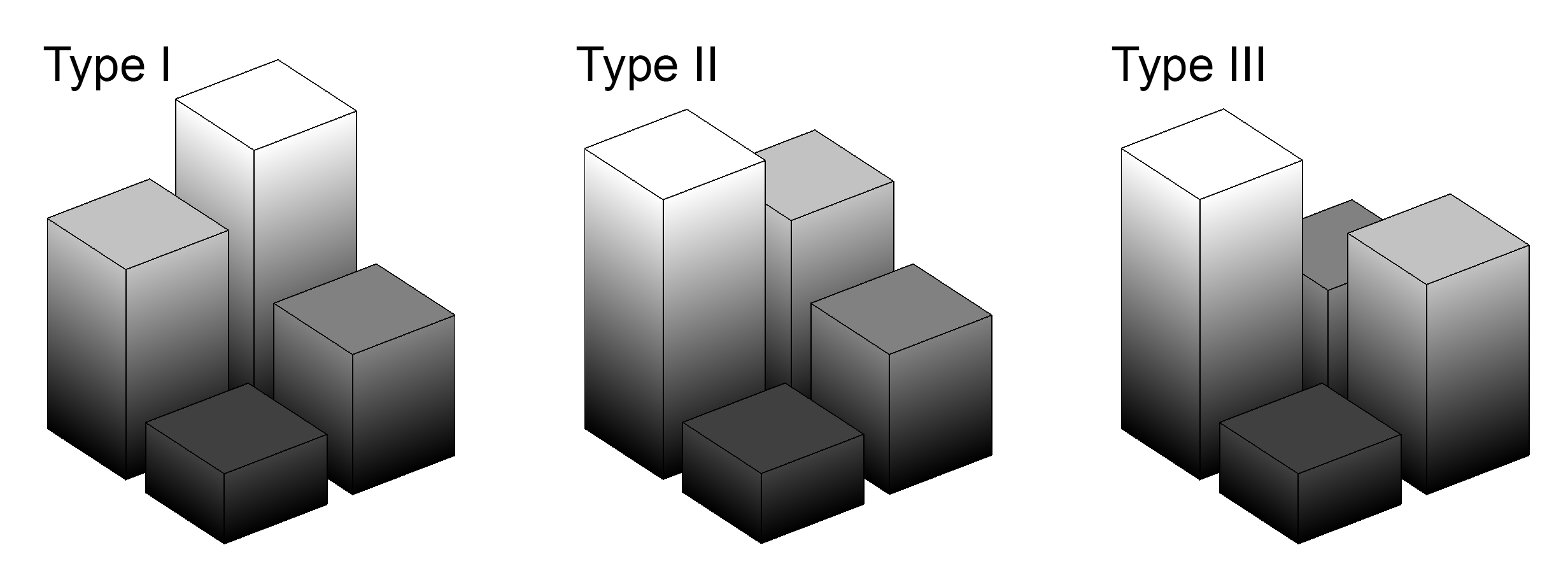}
\caption{\label{fig:3types} : Spatial visualization of the different types of ordinal patterns.}
\end{figure}

Figure~\ref{fig:3types} shows the spatial visualization of the three different types of ordinal patterns. In type I, values are either increasing in both rows or decreasing in both rows, and the same for columns. This type represents smoothness when gray values represent a function over a plane region, for example temperature. 

\begin{Proposition} Suppose the gray values $z(x,y)$ of our image over an open region all lie on a plane $z=ax+by+c$ in 3-space with $a,b\not= 0.$ Then all $2\times 2$ patterns for grid points in this region are of type I. This also holds if $z(x,y)$ is a function with continuous and non-zero partial derivatives in this region. \end{Proposition}

This is easy to see since the assumption says that in $x$-direction, the gray values are either strictly increasing in the whole region, or strictly decreasing in the whole region. The same for the $y$-direction.

For type II, the parallel increase or decrease holds for either rows or columns. In the other direction, we have one increase and one decrease. In a smooth surface, that oculd only happen where one partial derivative becomes zero somewhere inside our small square.  Of course images are not smooth. Type II frequently occurs at edges, curves and tree-like structures of an image.

In type III, both values of one diagonal are larger than both values of the other diagonal. In a smooth context, this is the rare case of a saddle point inside the little square. This type represents pure noise. For a checkerboard image, all $2\times 2$ patterns are of type III. For white noise, all three types occur with the same frequency. 

\begin{Proposition} The type of a $2\times 2$ pattern  is the rank number which shares a diagonal with 4. For instance we have type II if 2 is on one diagonal with 4. \end{Proposition}

This can be seen from Figure \ref{fig:ordPat2D_overview}. However, diagonal comparisons are not needed to decide about the type. Here is a little Matlab function {\tt t=ty(w)} which calculates the type directly from the local data $w=(w1,w2,w3,w4)$ in equation \eqref{locpat}.               

\begin{center} \tt  a=(w1$<$w2)+(w3$<$w4); if a==2; a=0; end\\
b=(w1$<$w3)+(w2$<$w4); if b==2; b=0; end\\
t=a+b+1; \end{center}

The concept of type of a $2 \times 2$ pattern seems new. However, there is a similar concept in statistics when we consider the interaction of two dichotomic variables  $x,y$ (`smoker' and `drinker', say) on a numeric variable $z$ (`blood pressure'). For a  $2 \times 2$ table of means of $z$ depending on $x$-$y$-combinations, the classification of interaction into  ‘pure ordinal’, ‘hybrid’, and ‘pure disordinal’ introduced by Leigh and Kinnear \cite{Leigh1980} exactly corresponds to our types I, II, and III.

\subsection{Frequencies and relevance of types}
Letting the above function run over all $2\times 2$ patterns \eqref{locpat} of our $M\times N$ data matrix $X,$ we obtain an $(M-1)\times (N-1)$ matrix $T$ of types. Then we determine the relative frequencies $q_1, q_2, q_3$ of 1,2,3 in $T.$ The same is done for the 24 patterns. See Ribeiro et al. \cite{Ribeiro2012}. They used the pattern probabilities $p=(p_1,...,p_{24})$ to define the standardized permutation entropy 
\begin{equation} S(p)=\frac{1} {\log 24} \sum -p_k \log p_k \ ,
\label{PE} \end{equation}
the Jensen-Shannon divergence $Q$ between $p$ and the equilibrium measure $p_e=(\frac{1}{24},...,\frac{1}{24}),$ and the complexity 
\begin{equation} C(p)= S(p) Q \quad\mbox{ with }\quad Q=\frac{1}{Q_{\rm max}} \left\{ S\left(\frac{p+p_e}{2}\right)-\frac12 S(p) -\frac12 S(p_e)\right\} \ ,
\label{Jcomp} \end{equation}
where $Q_{\rm max}= \frac12 (\log 96 -\frac{25}{24}\log 25)\ .$

In this paper we study an image database of Kylberg and Sintorn \cite{Kylberg2016, KSdatabase, Kylberg2014} which contains 100 samples for each of 25 textures and each of 9 rotations, altogether more than 20000 images of $122\times 122$ pixels. Details are given below. Here we explain why it makes sense to go from patterns to types.

\begin{figure}
\includegraphics[width=0.8\textwidth]{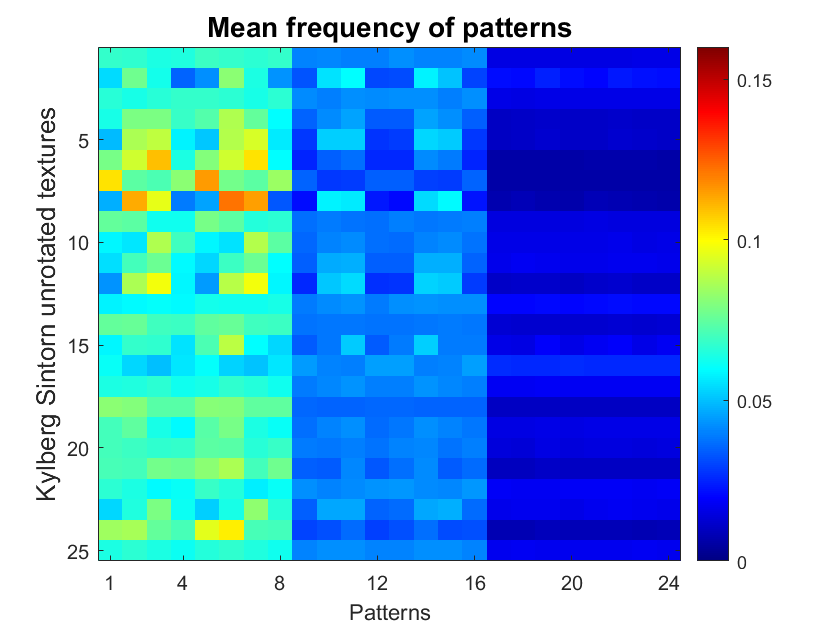}
\caption{\label{fig:OrdPat2D_mean} : Mean frequencies of patterns for 25 textures of the Kylberg Sintorn database. Patterns 1-8 are of type I, 9-16 of type II, and 17-24 of type III. Patterns of the same type occur with similar frequency.  Patterns of type I are most frequent, patterns of type III are rare.}
\end{figure}

Patterns were enumerated according to the three blocks of Figure \ref{fig:ordPat2D_overview}, and rowwise within each block. So patterns 1-8 are of type I, 9-16 of type II, and 17-24 of type III.  The rows in Figure \ref{fig:OrdPat2D_mean} correspond to 25 textures in the database. The color represents the mean frequency $p_k$ of patterns in 100 unrotated samples of each texture.

The picture shows a clear distinction between patterns of different types, throughout all textures.  The probability of a pattern is about 0.07 for type I, about 0.04 for type II and almost constant 0.015 for type III.  In the structures of rows 1, 3, 13, 17, 18, 20, 22 and 25,  the frequencies of patterns of the same type almost coincide. These are isotropic textures, as explained below. In such cases all the information of the pattern distribution $p$ lies in the probabilities $q_1,q_2,q_3$ of the types.

There are other images, rows 2, 5-8, 11, 12, 15 and 24, where probabilties within type I and/or II show clear differences. These textures have some dominating directions, and differences of the $p_k$ within types can give information about these directions. But the main part of the information of $p$ even in this case is contained in $q_1,q_2,q_3.$ From a practical viewpoint, the focus on types is justified. Compressing 24 probabilities $p_k$ to just three $q_j$ is connected with a small loss of information.

\subsection{The two parameters}
The type frequencies $q_j$ fulfil the equation $q_1+q_2+q_3=1 \, .$  So they are dependent and negatively correlated. It is better to select two parameters which are more independent. We define
\begin{equation} \tau = q_1 - 1/3 \quad \mbox{ and }\quad  \kappa = q_2-q_3 \ .
\label{para} \end{equation}
Both parameters are zero for white noise, that is, an array of independent random numbers with the same distribution. Moreover, the vectors of weight coefficients $(\frac23 , -\frac13 , -\frac13)$ for $\tau$ and $(0,1,-1)$ for $\kappa$ and the constant vector $(1,1,1)$ are orthogonal, in accordance with a standard principle for weight coefficients of contrasts in the analysis of variance. For one-dimensional ordinal patterns, similar parameters are discussed in \cite{Bandt2022}.

The parameter $\tau$ measures smoothness of the image and is defined in the same way as persistence $\tau$ of one-dimensional order patterns \cite{BandtShiha2007, Bandt2022}. Its maximum $2/3$ is assumed for smooth surfaces described in Proposition 1. The minimum is $-1/3$ but $\tau$ rarely assumes negative values.  The parameter $\kappa$ is more difficult to interpret. In some way it describes how much curve and tree structures dominate checkerboard-like noise in the image. See the examples in Section \ref{appli}. The theoretical bounds of $\kappa$ are -1 and 1, its range in real data is between -0.1 and 0.5.

\subsection{Discussion of dependence and correlation}
For one-dimensional patterns, dependence of successive patterns is a big problem which makes multinomial models for pattern frequencies invalid, see Elsinger \cite{Elsinger2010}, Wei{\ss} \cite{Weiss2022}, de Sousa and Hlinka \cite{Sousa2022}. For $2\times 2$ neighboring patterns, transition probabilities  were investigated and corresponding Markov chains constructed by Pessa et al. \cite{Pessa2020}. Horizontal and vertical neighbors have to be distinguished. Many pairs of patterns cannot occur as neighbors. This creates strong statistical dependence between pattern probabilities $p_k.$

Fortunately, the dependence between our parameters is much weaker.  There are no forbidden pairs of neighbor types. To check the dependence more carefully, we consider the statistics of types in a $3\times 3$ square. Any two neighbor patterns are within such a $3\times 3$ block. The 4 types at the upper left, upper right, lower left and lower right within such a block can be quickly calculated for all possible $9!=362880$ permutations. We obtain a type matrix $M$ with $9!$ rows and 4 columns. Treating all permutations equally means that we assume white noise as model for the $3\times 3$ data matrix. 

Our calculation showed that the types of diagonal neighbors are uncorrelated, which also follows from symmetry arguments. For two neighboring $2\times 2$ patterns in a row or column, let $Y_j=1$ if the first pattern has type $j,$ and $Y_j=0$ otherwise. Note that $Y_j$ has mean $1/3$ and variance $2/9.$ Let $Z_k=1$ if the neighbor has type $k$ and $Z_k=0$ otherwise. Matlab evaluation of \ {\tt mean(M(:,1)==j \& M(:,2)==k)-1/9} \ for the long type matrix $M$ gave the neighbor covariance matrix $C$ with $c_{jk}={\rm Cov}(Y_j,Z_k).$
\[ C=(c_{jk}) = \frac{1}{180}\cdot\left(\begin{matrix}
1&0&-1\\ 0&1&-1\\-1&-1&2
\end{matrix}\right) \ . \]

Now take the $(M-1)\times (N-1)$ type matrix $T$ obtained from an $M\times N$ data matrix. Let $U=(M-1)(N-1),$ and let $Q_j$ be the number of entries $j$ in matrix $T.$ We have  $Q_j=\sum_{u=1}^U Y^u_j$  where $u$ runs through the places $(m,n)$ in the matrix and $Y^u_j=1$ if the entry in $T$ at place $u$ is $j,$ and 0 otherwise. By the linearity of covariance we have
\[ {\rm Cov}(Q_j,Q_k)=\sum_{u=1}^U \sum_{v=1}^U {\rm Cov}(Y^u_j,Y^v_k) \ .\]
Under the assumption of white noise, this sum contains zeros for all pairs of places $u,v$ which are not equal and not neighbors in a row or column - since the types of such pairs of places are independent. We are left with $U$ terms for which $u=v,$ and $4U$ terms for which $v$ is one of the 4 neighbors of $u.$ Actually we have less neighbors if $u$ is in the first or last row or column. This border effect is neglected since we look for an asymptotic formula.

For the $4U$ neighbor cases the covariance $c_{jk}$ is taken from the above matrix $C.$ For the $U$ cases with $u=v$ we have the variance $2/9$ when $j=k,$ and the covariance $0-(1/3)^2=-1/9$ when $j\not= k.$ We get
\[ {\rm Cov}(Q_j,Q_k)=U\cdot(4 c_{jk}+2/9)\ \mbox{ if }j=k\quad
\mbox{ and } \quad U\cdot(4 c_{jk}-1/9)\ \mbox{ else. } \]
If we consider relative frequencies $q_j=Q_j/U$ we have  $U\, {\rm Cov}(q_j,q_k)= \frac{1}{U} {\rm Cov}(Q_j,Q_k)\, .$ 

\begin{Proposition}  Under the assumption that the random image is white noise, and neglecting the border effect, we get the asymptotic formula
\[ U\cdot {\rm Cov}(q_j,q_k) = \frac{1}{45}\left(\begin{matrix}
11&-5&-6\\ -5&11&-6\\ -6&-6&12
\end{matrix}\right) \ . \]
Thus  ${\rm Var\,} \tau =\frac{11}{45}, {\rm Var\,} \kappa ={\rm Var\,}q_2+{\rm Var\,}q_3 -2{\rm Cov}(q_2,q_3)=\frac{35}{45}\, ,$ and ${\rm Cov\,} (\tau ,\kappa) =\frac{1}{45}.$  Thus the correlation coefficient of  $\tau$ and $\kappa$ is $1/\sqrt{385} \approx 0.05.$
\end{Proposition}

In other words, $\tau$ and $\kappa$ are almost uncorrelated, in contrast to pattern frequencies in the one- and two-dimensional case \cite{Elsinger2010, Pessa2020, Weiss2022, Sousa2022}.  The calculation also revealed that the parameters  $q_3-1/3$ and $q_1-q_2$ are completely uncorrelated.  However, these parameters did not classify textures well. Their values were in a narrow strip, like the entropy-complexity combinations in Figure \ref{fig:fBM_newPar12_ECPlane}.  Favouring classification strength more than total lack of correlation, we decided to take $\tau$ and $\kappa$ as our parameters.

\begin{figure} 
\includegraphics[width=0.6\textwidth]{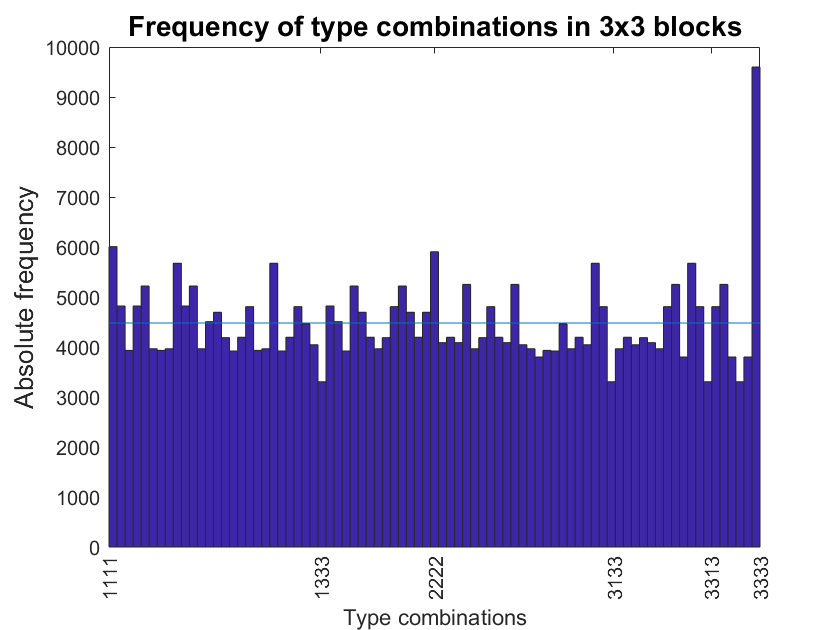}
\caption{\label{3depen} : Histogram of the 81 possible combinations of the 4 types in a $3\times 3$ block, lexikographically ordered from 1,1,1,1 up to 3,3,3,3. Only the combination of four types III strongly exceeds the mean frequency indicated by the line. Other multidimensional dependencies seem to be weak.}
\end{figure}

For two-dimensional patterns, uncorrelated $q_k$ need not be almost independent since central limit theorems do not apply. For that reason we checked the combinations of 4 types in a $3\times 3$ matrix, using the above mentioned long type matrix $M$ for all $9!$ possible permutations. It turned out that indeed the probability of type III increases from 0.333 to 0.367, 0.414 and 0.575 if we assume that one, two in a line, or three neighboring patterns within the $3\times 3$ block are type III. However, the histogram of all possible combinations of the 4 types shows for 3,3,3,3 the only frequency which considerably exceeds the mean $9!/81,$ indicated by the line in Figure \ref{3depen}. As compensation, the four combinations of one type I and three types III have the smallest frequency. Combinations 1,1,1,1 and 2,2,2,2 have second highest frequency. Altogether, dependencies caused by two or three neighbors apparently do not change the picture.

\section{Applications and examples}\label{appli}

\subsection{The Kylberg Sintorn rotation dataset of textures}
The Kylberg Sintorn rotation dataset consists of $25$ different textures obtained from  bulk solids like lentils, grains and sprinkles as well as more regular structures like woven materials and knitwear. The database is publicly available at \cite{KSdatabase}.
Each texture and rotation is represented by $100$ image samples, which were obtained by cutting the original photo into small tiles following a $10$ by $10$ grid. Each tile has a size of $122\times 122$ pixels and the gray values were normalized to a mean value of $127$ and a standard deviation of $40$. Pictures of all 25 textures can be found at \cite{KSdatabase} as well as in the paper of Kylberg and Sintorn~\cite{Kylberg2016} and the dissertation \cite{Kylberg2014}. Figure \ref{KylbergPhotos} below shows six of the structures.

All photos are also provided under 8 rotations by multiples of $40^{\circ}.$ 
Rotations of the textures were produced with different techniques. Part of the data comprises software rotations of the textures using different interpolation methods, and the goal of research was the comparison of such algorithms. The highlight of the database are hardware rotations, implemented by turning the camera. Figure 2 in \cite{Kylberg2016} shows the sophisticated setup to derive photos of comparable quality for all textures and angles. In our work, we only use these hardware rotations. Thus we have 900 samples of each structure, 100 unrotated and 800 rotated ones. 

\begin{figure}
\includegraphics[width=0.95\textwidth]{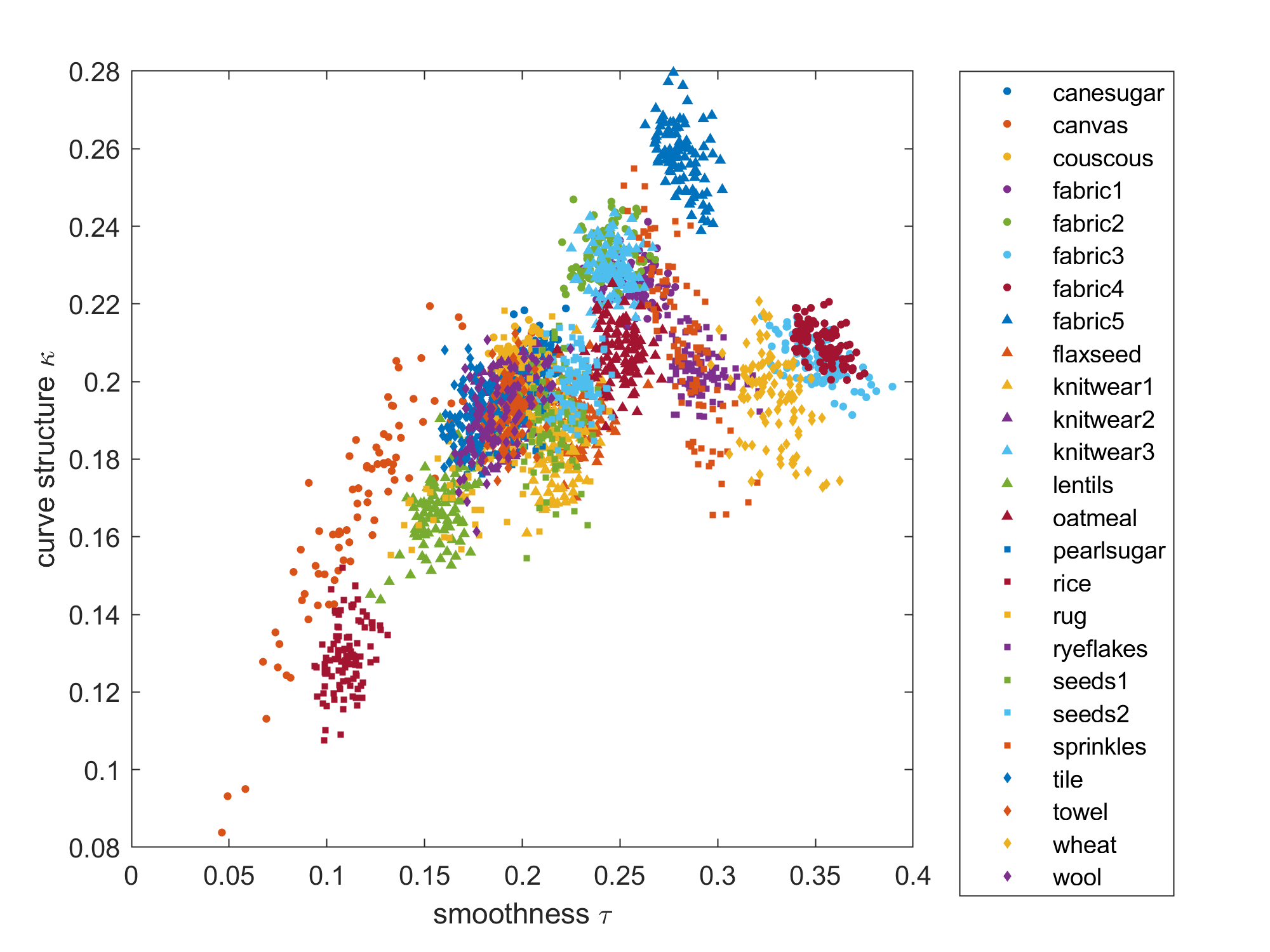}
\caption{\label{Kylberg_newPar} : Description of the 25 textures from the Kylberg Sintorn database by our parameters. The samples of each texture are represented by 100 points.}
\end{figure}

\subsection{Results for unrotated textures}
Figure~\ref{Kylberg_newPar} shows the parameter pairs $(\tau ,\kappa)$ for all 2500 unrotated samples of Kylberg and Sintorn. Each texture is represented by a cluster of 100 points. 
It is surprising how small and compact the clusters are, with exception of sprinkels, canvas, rug, and wheat. Although the images are fairly small, the parameters $\tau$ and $\kappa$ are consistently estimated with differences of less than $\pm 0.02$ to the cluster mean.

\begin{figure}
\includegraphics[width=0.8\textwidth]{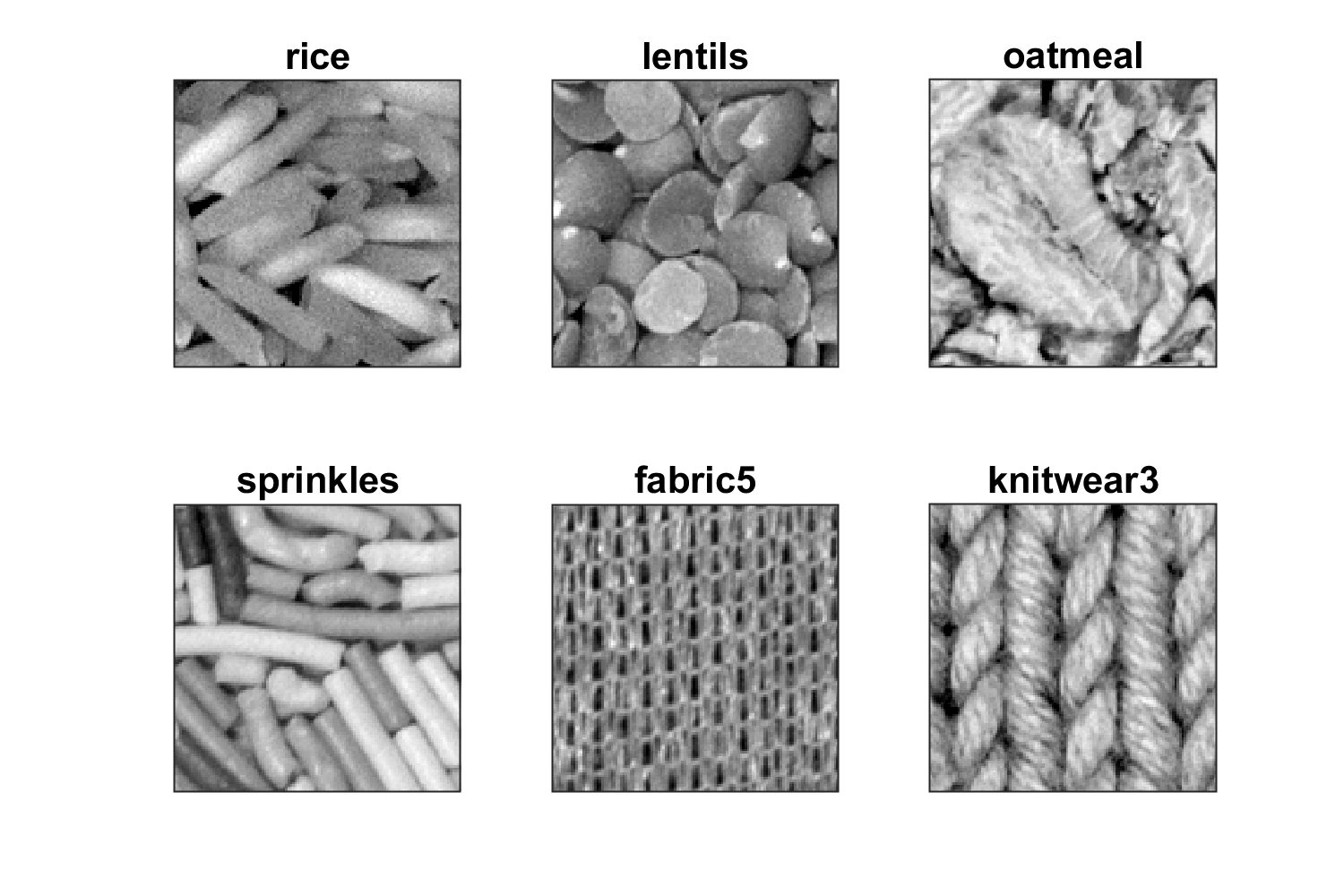}
\caption{\label{KylbergPhotos} : Six textures from the Kylberg Sintorn database. Their mean pattern frequencies are found in rows 16, 13, 14 and 21, 8, 10 of Figure \ref{fig:OrdPat2D_mean}.}
\end{figure}

Some of the materials are clearly identified by the two parameters. Rice and lentils are separated from all types of fabric and knitwear. They have many patterns of type III and rather few of type I, resulting in small values of $\tau$ and $\kappa .$ In case of rice, this seems partly due to noise in the photo, see Figure \ref{KylbergPhotos}. Figure \ref{fig:OrdPat2D_mean} shows that for rice depicted in row $16$ all type I patterns have small frequencies. Some  noise, from the camera sensor or from our treatment of equal values, can also be seen in lentils. Of course our parameters do not allow to distinguish effects of the material, light sources, or camera focus settings.

The largest smoothness $\tau$ was obtained for fabric 3 and 4 which have almost the same parameter values. The largest $\kappa$ is reached by fabric 5 which has many curves almost parallel to the axes, cf. Figure \ref{KylbergPhotos}. This value will decrease when the material is rotated. Already in Figure \ref{fig:OrdPat2D_mean}, fabric 5 in row 8 showed different pattern probabilities within type I, which indicates dominating directions in the texture.

Finally, let us compare rice and sprinkels in Figure \ref{KylbergPhotos}. Both materials consist of lengthy kind of cylinders. However, rice grains are shorter and arranged in an irregular way while sprinkels are longer and form bundles of parallel pieces. This leads to preferential directions in the samples although there are no preferred directions in the large photo. That seems the reason for the large and irregular size of the sprinkels cluster in Figure \ref{Kylberg_newPar}. Actually, also the frequencies of patterns of type I did vary much more for sprinkels than for any other texture.

The results of this experiment were encouraging, but must not be overestimated. There are problems with equal gray values which affect up to 70 percent of the $2\times 2$ patterns. The method of adding noise could in such cases create artificial noise. It is a surprise that it works so well.

\begin{figure}
\includegraphics[width=0.8\textwidth]{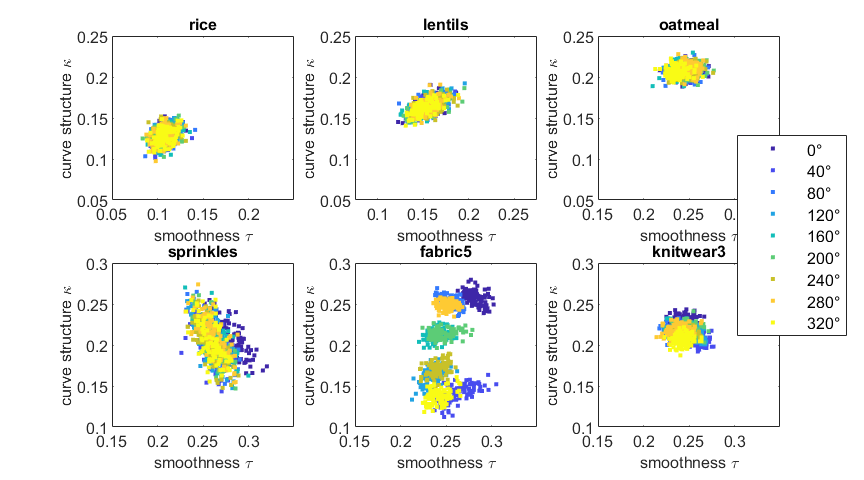}
\caption{\label{fig:Kylberg_rotations} : Parameter pairs for all 900 rotated samples of each of the textures of Figure \ref{KylbergPhotos}. The upper row shows that for isotropic textures there are no rotation effects.  In case of fabric 5, horizontal and vertical line structures cause drastic changes of the parameters during rotation.}
\end{figure}

\subsection{Results for rotated textures}
For the six textures in Figure \ref{KylbergPhotos} we now consider the parameters of the rotated photos. Together with the unrotated samples, we plot 900 points for each material.
It turns out that for isotropic textures, as shown in the upper row of Figure \ref{KylbergPhotos}, there is no effect of rotation. Figure \ref{fig:Kylberg_rotations} shows that clusters stay in the same location. Although our parameters are based on square patterns, they are rotation-invariant for isotropic textures. Note that the clusters for rice, lentils, and oatmeal are disjoint, according to change of the $x$-scale.

The non-isotropic textures in the lower row behave differently. Knitwear 3 has clear vertical structures which apparently did not interact with rotations.  Our parameters seem to evaluate the irregularly directed mini-threads in the photo. \ 
The long cylinders of sprinkles already caused rotation effects among the samples in the unrotated case, as discussed above. Rotating the whole picture did not much increase this variation. 
However, for some of the materials we noticed a small shift of one rotation to higher $\tau$ values, e.g. the dark blue squares for sprinkles in Figure~\ref{fig:Kylberg_rotations}. A reason may be that the affected photos were taken with a slight change of focus, which resulted in a softer and smoother image. 

Fabric 5 shows very clear rotation effects, which were also seen in other fabrics and canvas.
This is due to the clear horizontal and vertical structures.  Unrotated pictures show very large $\kappa$ values, cf. Figure \ref{Kylberg_newPar}.  Rotations around $40^o$ and $320^o$ are associated with the smallest $\kappa ,$ followed by $120^o$ and $240^o.$ Nearest to the unrotated values are rotations by $80^o$ and $280^o,$ followed by $160^o$ and $200^o.$ Thus we expect the smallest change for multiples of $90^o$ and the largest effect for $45^o.$ 

\begin{figure}
\includegraphics[width=0.8\textwidth]{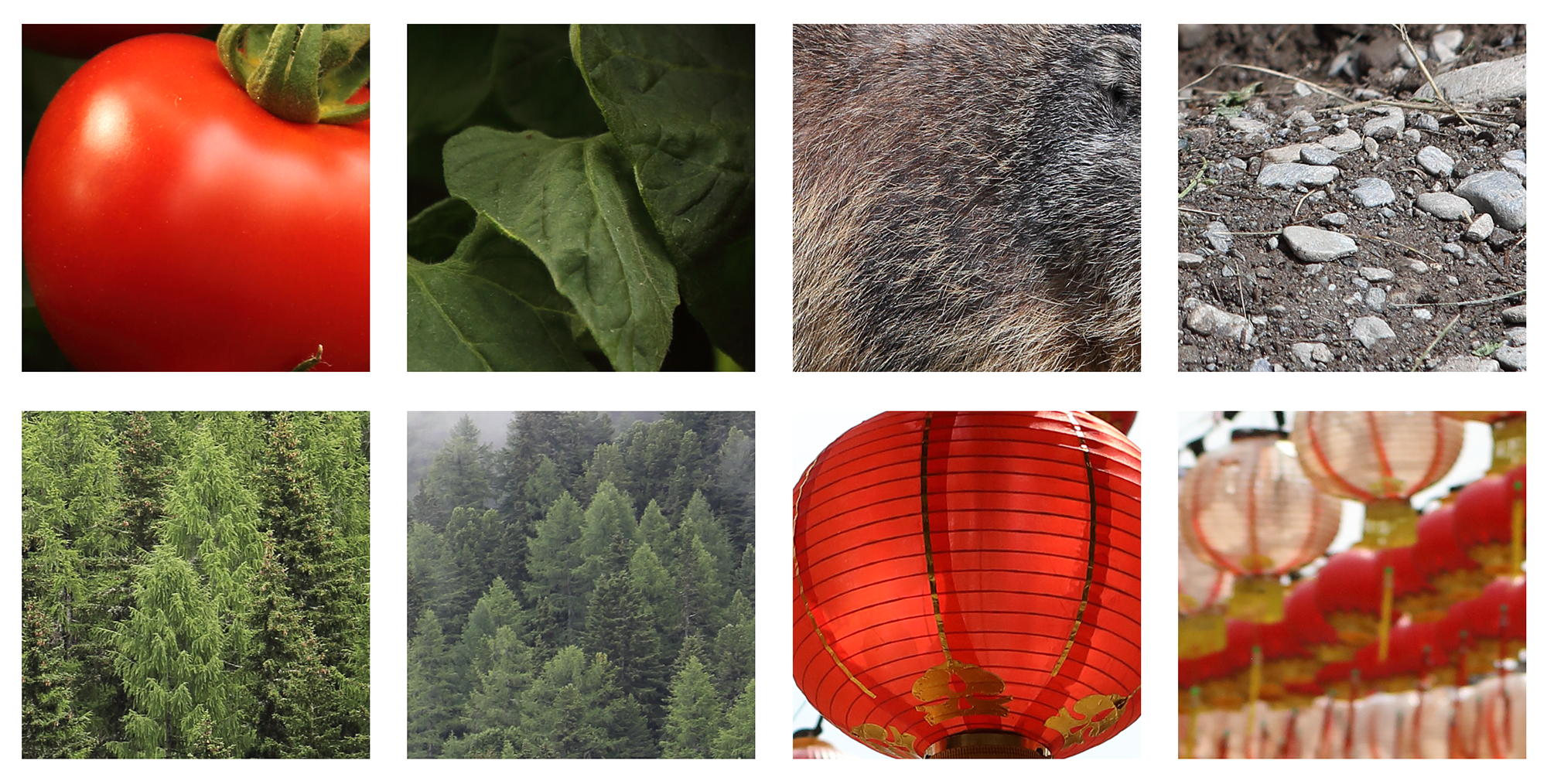}\vspace{2ex}\\
\includegraphics[width=0.9\textwidth]{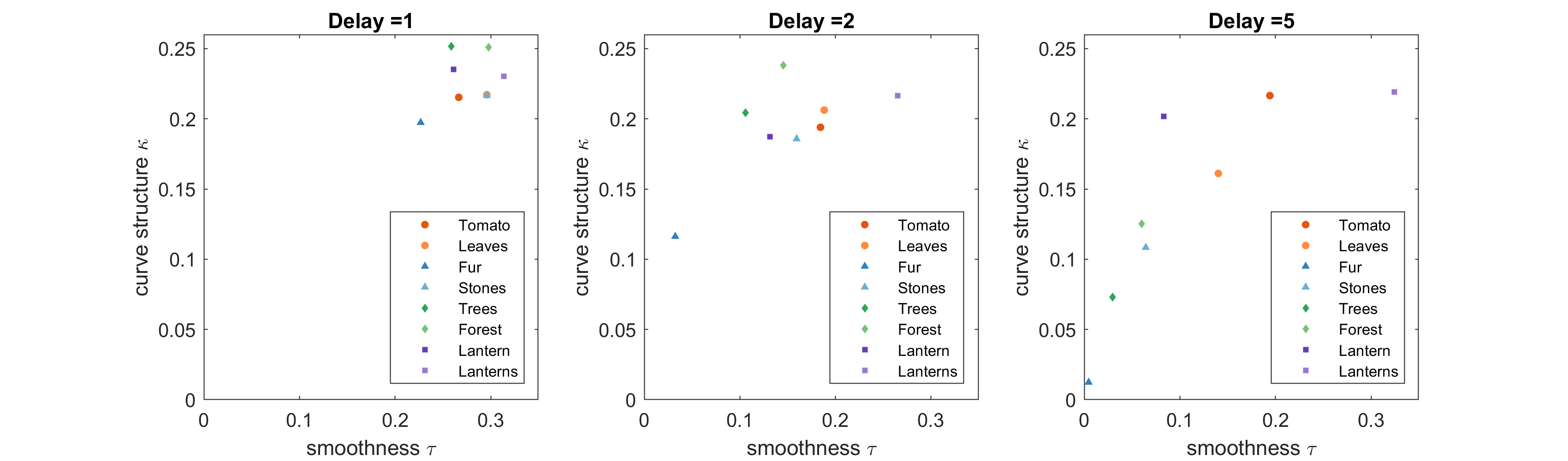}
\caption{\label{photos8} : Four pairs of details from photos and their parameters for delay 1, 2 and 5.}
\end{figure}

\subsection{Photos}
Can we test our parameters also with our own archive of photos? There are some difficulties.  The pixel structure can only be evaluated if we have a pixel format, like tiff, png or bmp. The compressed jpeg format is not appropriate.  Next, color photos must be transformed to grayscale images.
For this step, there are different possibilities to convert an RGB image to a grayscale image and extract the luminance from the three color channels (R=red, G=green, B=blue). We have decided to use the following formular:
\[I = 0.299*R + 0.587*G + 0.114*B\]
which is widely used in academics and implemented in the MATLAB function \texttt{rgb2gray}.

A main problem is the high resolution of photos compared to low resolution of graytones which causes a lot of equality among values of neighboring pixels. Moreover, photos usually contain many objects with further details so that there is no uniform texture at all. In order not to get a mean of many different textures, one has to consider small parts of the photo.

Figure \ref{photos8} shows four pairs of detail photos with $640\times 640$ pixels cut out from four larger photos. The delay $d=1$ used for the textures does not give much difference between the parameters of the photos, due to little difference of values of neighboring pixels. On this scale, all photos were rather smooth. Only for delay 5 the character of the image becomes visible. Lanterns and tomato remain smooth while the fur of an animal has parameters near $(0,0),$ the point of white noise. Fur and trees with their fractal structure involve a lot of type III patterns. Forest was taken from the background with some fog which smoothes the structure.

\subsection{Fractal surfaces}
Fractal surfaces are one of the few model classes of images which can be generated by a stochastic algorithm. Depending on the Hurst exponent $H\in [0,1]$, the resulting surface is rough for small $H$ and smooth for larger $H$. Ribeiro et al. \cite{Ribeiro2012} and Zunino and Ribeiro \cite{Zunino2016} studied ordinal patterns in fractal surfaces and represented them in the entropy-complexity causality plane.
For our study we also used the midpoint displacement algorithm as described by Peitgen and Saupe \cite{Barnsley1988}. We simulated 100 fractal surfaces of size $2049\times 2049$ for each Hurst exponent $H=0.1, 0.2, \ldots, 0.9\, .$ We determined the distributions of  $2\times 2$ patterns and types with embedding delays $d=1,\ldots,200$. Based on these distributions, we calculated entropy and complexity in accordance with \cite{Ribeiro2012, Zunino2016}, and our two parameters. 

\begin{figure}
\includegraphics[width=0.9\textwidth]{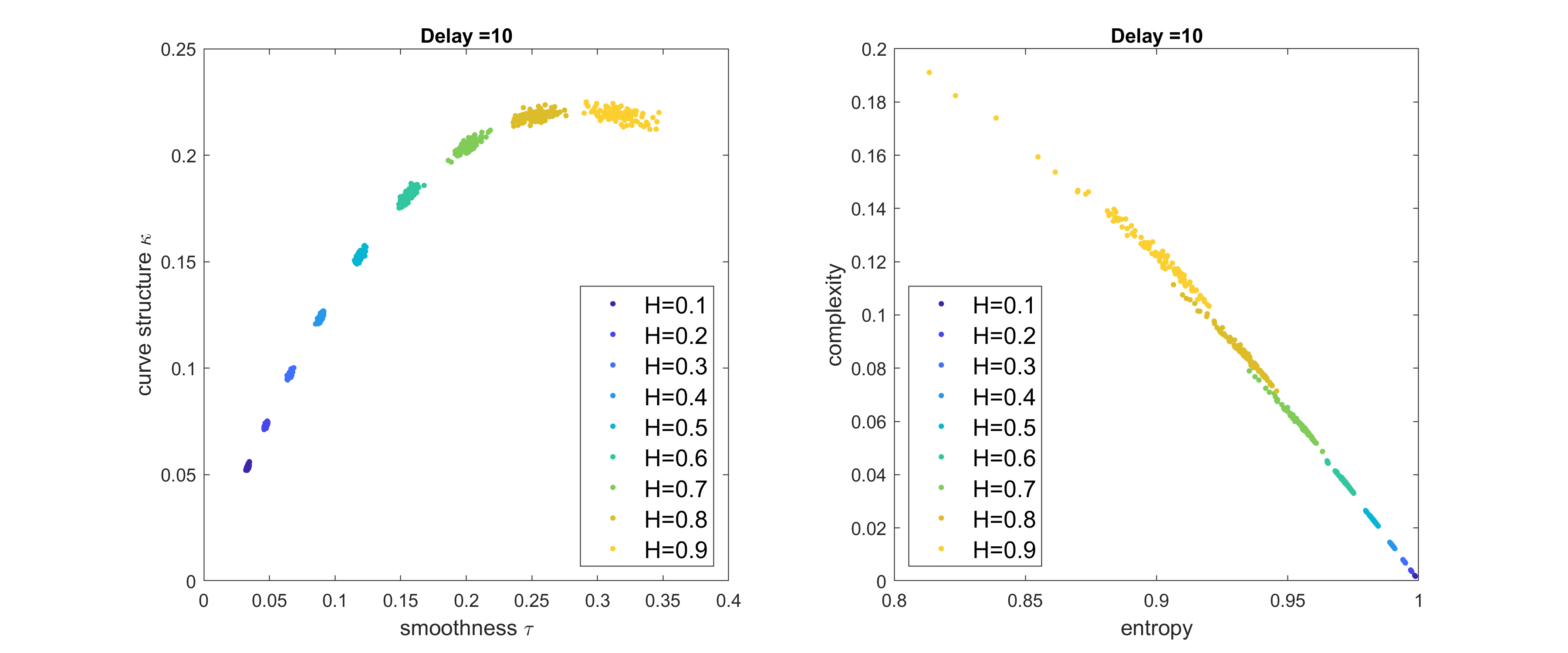}
\caption{\label{fig:fBM_newPar12_ECPlane} : Extracted features from 100 fractal surfaces for each Hurst exponent $H=0.1,...,0.9$. Parameters $\tau$ and $\kappa$ (left) and entropy-complexity plane (right), cf. \cite{Ribeiro2012, Zunino2016}}
\end{figure}

Due to the self-similarity of fractional Brownian motion, the results do not essentially depend on $d,$ see \cite{Zunino2016}. However, there are numeric effects for small and large $d.$  The smallest variation was obtained for $d=10$ which is very similar to 20 or 50. Figure \ref{fig:fBM_newPar12_ECPlane} shows separated point clouds for the parameter values. As expected, large $H$ corresponds to large values of $\tau$ and $\kappa$ indicating smoothness. For small $H,$ the point of white noise (formally $H=0$) is approached, which is $(0,0)$ for our parameters and $(1,0)$ in the entropy-complexity plane.

\section{Conclusion}
We introduced two very simple ordinal parameters describing smoothness and curve structure in images. They could become part of the big toolbox of image processing. Applications range from virus detection to the analysis of satellite images. Our parameters were tested with the Kylberg Sintorn rotation database. They showed small variation in samples of the same texture and proved to be amazingly invariant under rotations for isotropic textures. Many structures can be separated by using just these two parameters.

Of course the study of $2\times 2$ patterns expresses specific features of the microstructure of images. Larger patterns will provide more information.  A principal obstacle is the number of possible permutations which vastly increases with the size of pattern.  Our study shows, however, that by grouping into meaningful types this number can be drastically reduced.

\section{Acknowledgments}
This paper is dedicated to Karsten Keller. We gratefully remember the time when we cooperated with him. With his great exploratory spirit and educational experience, he was a wonderful advisor for the diploma thesis of KW.

\section{Author declarations}
The authors have no conflicts to disclose.

\section{Data availability}
The Kylberg Sintorn rotation dataset is publicly available.~\cite{KSdatabase} \ 
The eight photo tiles extracted from photos of KW are available from the corresponding author.

\end{document}